# Pre-processing of Domain Ontology Graph Generation System in Punjabi

Rajveer Kaur[#1], Saurabh Sharma[*2]

[#]*Research Scholar,* [*]*Assistant Professor*

[#*]*Department of Computer Science, Baddi University of Emerging Sciences and Technology, Baddi, India*

*Abstract-* **This paper describes pre-processing phase of ontology graph generation system from Punjabi text documents of different domains. This research paper focuses on pre-processing of Punjabi text documents. Pre-processing is structured representation of the input text. Pre-processing of ontology graph generation includes allowing input restrictions to the text, removal of special symbols and punctuation marks, removal of duplicate terms, removal of stop words, extract terms by matching input terms with dictionary and gazetteer lists terms.**

## I. INTRODUCTION

"Ontology" has originated from Philosophy branch but in the last decades, Ontology in information systems has become common in numerous other fields like Natural Language Processing, Database integration, Internet technologies, Artificial Intelligence, Multiagent Systems etc.

Informally Ontology describes the terms, concepts, classes and relationship between them. Formally Ontology is defined as: "formal, explicit specification of a shared conceptualization". Conceptualization [4] is an abstract view of the world that represents the knowledge for any motive. Explicit implies that the variety of concepts used and therefore the constraints on their utilization are explicitly outlined. Formal make reference to the reality that ontology ought to be machine understandable and readable. Shared reveals the insight that an ontology captures accordant knowledge i.e. it's not individual personal information, however accepted by a group.

The ontology graph is a new methodology that is employed to represent the ontology of knowledge in a domain. The ontology graph comprises of diverse levels of conceptual components, in which they are related together by various types of relations. It is essentially a dictionary framework (i.e. terms) that unite one another to represent a class, to formulate various concepts and identify meanings. The conceptual layout of an ontology graph comprises of numerous terms with a few relationships between them, so that the different conceptual components are framed like a lattice.

An ontology graph modeling process includes two phases: Pre-processing phase and Processing phase. The aim of this paper is to represent the pre-processing phase of domain ontology graph generation system for Punjabi text documents.

Punjabi language is a part of Indo-Aryan language. Punjabi language is the formal language of the Punjab, one of the states of India. Punjabi is spoken in Punjab, Punjab Province of Pakistan, and Jammu-Kashmir. It's conjointly use for communication as a minority language in several other countries where Punjabi people are living, such as the United States of America, Australia, Canada, and England. Punjabi is written in the two different scripts; Gurumukhi script in India and Shahmukhi script in Pakistan.

The Punjabi text document corpus contains 1000 documents with an average of 700 Punjabi words in each document. The text documents belong to five different domains, which are Agriculture, Entertainment, Health, Politics, and Sports. Each domain contains 200 Punjabi documents. These five domains are labeled as classes for the domain ontology learning process. The documents of the corpus in every class are further divided to permit 70% of them for the learning set and 30% for the testing set.

## II. PRE-PROCESSING PHASE

Text Preprocessing is an extremely important part of any Natural Language Processing system, since the characters, terms recognized at this stage are the basic units goes to further processing stages. Pre-processing phase represents the input text in organized manner. In pre-processing phase, manipulation and filtration of terms is carried out to omit/eliminate terms that do not consist of context like stop words, special symbols, punctuation marks etc. When Pre-processing has been done, only those terms are remaining that are meaningful to the Natural Language Processing System. Pre-processing phase includes: Allowing input restrictions to input text, elimination of useless symbols, removal of duplicate terms, Removal of Stop words, Extract terms matched with dictionary terms, and Extract terms matched with Gazetteer list.

Term Extraction is a pre-processing phase that identified all meaningful Punjabi terms in input text documents. An existing electronic dictionary is utilized. It contains 40,000 Punjabi words. It is very helpful for identifying the meaningful terms inside the input text. To extract the terms, besides the existing dictionary terms, an additional input of Punjabi terms are also required. These additional terms are Middle names, last names, Politicians names, Sports personality names, Sports names, Movies names, Location names etc. To extract the meaningful terms and remove the meaningless terms, Stop Word list is required. We created

Stop list manually by analyzing Punjabi documents. The Stop Word List contains 1,500 words.

The architecture of Pre-processing phase is given in Fig. 1.

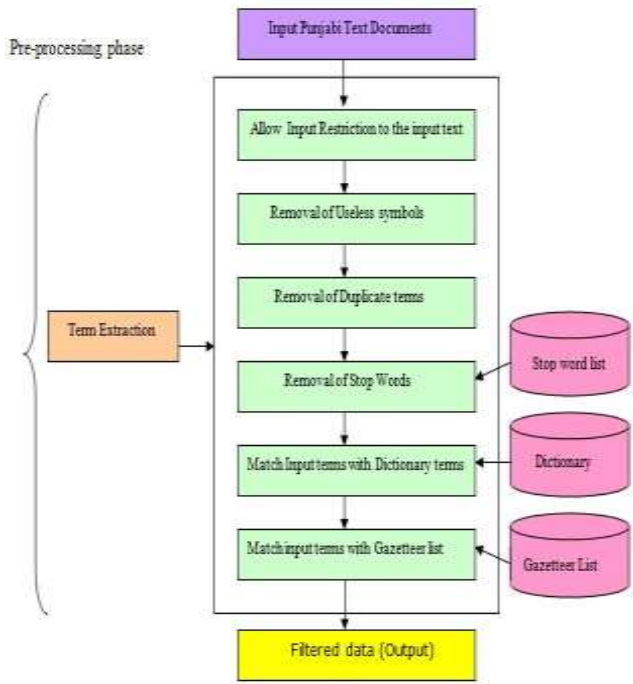

Fig. 1 Architecture of pre-processing phase of ontology graph modeling

Various sub phases of Term Extraction are as follows:

*A. Allowing Input Restrictions to the input text*

The proposed system allows only Unicode based text as input i.e. it allows only Gurumukhi text as input. In Punjab, the standard for writing the Punjabi language is Gurumukhi script. The range of the Unicode text is given. System checks first character, if it is lies between the Unicode character ranges then it accepts the input, otherwise it rejects the input and again checks next character and so on.

*B. Removal of Useless Symbols*

This phase of Pre-processing includes elimination of useless symbols that are meaningless. Types of useless symbols are:
   a. Special Symbols and Punctuation marks (<,>, :,{,},[,],^,&,*,(,) etc.)
   b. Extra tabs
   c. Extra spaces
   d. Shift
   e. New line

System checks the input character one by one, if character is a special symbol or punctuation mark, skips the character and checks the next character. It only adds the Unicode character in the file. To remove extra white spaces and tabs, system checks the character with the previous character, if previous character matched with current character then it does not add this character in file and set this character as previous character and again check for next character i.e. if previous character and current character both are commas then only one comma is added to the file. If all terms in a file is in new line, then replace the new line with comma (,). After removal of symbols, output file contains only comma separated terms.

*C. Removal of Duplicate Terms*

System omit the duplicate terms from the input Punjabi text documents. Duplicate terms are omitted from input by comparing the terms. If duplicate term found then set this term as null otherwise it is added to the terms list. Removal of duplicate terms from input hinders duplicate terms from come into sight in final output. It is essential to omit duplicate terms from input text because it minimized the number of input terms and takes less time for further processing.

*D. Remove Stop Words*

High-frequency terms occurring in input text are well known as stop words. E.g. ਹੋਣ (hōṇ), ਦੇ (dē), ਨੂੰ (nūṃ), ਨਾਲ (nāl), ਉੱਥੇ (utthē), ਤੇ (tē) etc. These terms have to be discarding from input text because these terms do not have relevant information to the Punjabi ontology generation.

We have created Punjabi language Stop word list manually from Punjabi input data set. We manually analyzed the data set and identified 1,500 stop words.
Example:
**Input:** ਪੰਜਾਬ ਦੇ ਸਰਕਾਰੀ ਖੇਤਰ ਵਿੱਚ ਕੋਈ ਵੀ ਆਧੁਨਿਕ ਭੰਡਾਰ ਘਰ ਨਹੀਂ ਹੈ

Pañjāb dē sarkārī khētar vicc kōī vī ādhunik bhaṇḍār ghar nahīṃ hai

**Output:** ਪੰਜਾਬ ਸਰਕਾਰੀ ਖੇਤਰ ਆਧੁਨਿਕ ਭੰਡਾਰ ਘਰ

Pañjāb sarkārī khētar ādhunik bhaṇḍār ghar

Six stop words (ਦੇ (dē), ਵਿੱਚ (vicc), ਕੋਈ (kōī), ਵੀ (vī), ਨਹੀਂ (nahīṃ), ਹੈ (hai)) are removed from output sentence.

*E. Extract Terms by matching input terms with dictionary terms*

After removal of useless symbols, stop words, and duplicate terms; Punjabi Dictionary is used to match terms. It contains 40,000 words. Dictionary is used to identify the meaningful terms in input text. Terms are extracted by matching input terms one by one with dictionary terms. Terms that are matched with dictionary terms stored in accepted terms file. Remaining terms that are not matched with dictionary terms are stored in remaining term file for further processing.

*F. Extract terms by matching input terms with Gazetteer List*

Remaining input terms that are not matched with dictionary terms are further compare with Gazetteer list. Terms that are matched with Gazetteer list is stored in file where dictionary matched terms are stored. Terms that are not matched stored in rejected terms file. Punjabi Gazetteer list is created manually because no corpus is available on the web. Gazetteer list includes Middle names, last names, Politicians names,

Sports personality names, Sports names, Movies names, Location names etc. Gazetteer list contains 4,134 terms.

## III. ALGORITHM FOR PRE-PROCESSING OF TEXT DOCUMENTS

Following is the proposed pre-processing algorithm to build ontology graph from Punjabi language documents of five predefined classes such as Agriculture and Environment, Entertainment, Health, Politics, and Sports.

Step1: Apply input restrictions on input text i.e. only Unicode characters are accepted.

Step2: Remove useless symbols i.e. special symbols (<, >, :,{,},[,],^,&,*,(,) etc.), extra tabs, extra white spaces and shift from the Punjabi text documents.

Step3: Remove duplicate terms from the input text documents.

Step4: Remove Stop words e.g. ਦੇ (dē), ਵਿਚ (vicc), ਦੀ (dī), ਹੈ (hai), ਇਹ (ih), ਹਨ (han), ਨੂੰ (nūṃ) etc. using Punjabi Stop Word List.

Step5: Match input terms with dictionary terms to extract meaningful terms.

Step6: Match input terms with Gazetteer lists terms to extract names, places, location name, movies names etc.

Example of Pre-processing is shown in Table 1.

TABLE I
SAMPLE INPUT AND OUTPUT SENTENCES, FOR PREPROCESSING ALGORITHM.

| | |
|---|---|
| Sample Input | ਦੁਨੀਆਂ ਵਿੱਚ ਰੁਜ਼ਗਾਰ ਨਾਲ ਸਬੰਧਿਤ ਸਭ ਤੋਂ ਵੱਡੀ ਸਕੀਮ ਮਹਾਤਮਾ ਗਾਂਧੀ ਰਾਸ਼ਟਰੀ ਪੇਂਡੂ ਰੁਜ਼ਗਾਰ ਸਕੀਮ ("ਮਨਰੇਗਾ") ਪੰਜਾਬ ਵਿੱਚ ਸਿਆਸੀ ਅਤੇ ਪ੍ਰਸ਼ਾਸਨਿਕ ਬੇਰੁਖ਼ੀ ਦਾ ਸ਼ਿਕਾਰ ਦਿਖਾਈ ਦਿੰਦੀ ਹੈ। ਮਨਰੇਗਾ ਕਾਨੂੰਨ ਨਾਲ ਗ਼ਰੀਬ ਪਰਿਵਾਰਾਂ ਨੂੰ 100 ਦਿਨ ਦਾ ਰੁਜ਼ਗਾਰ ਪ੍ਰਾਪਤ ਕਰਨ ਦਾ ਸੰਵਿਧਾਨਕ ਹੱਕ ਮਿਲ ਗਿਆ ਸੀ। ਇਹ ਪਹਿਲੀ ਸਕੀਮ ਹੈ ਜੋ ਮੰਗ ਉੱਤੇ ਆਧਾਰਿਤ ਹੈ, ਕਿਸੇ ਸਰਕਾਰ ਜਾਂ ਅਧਿਕਾਰੀਆਂ ਦੀ ਗਰਾਂਟ ਉੱਤੇ ਨਹੀਂ। <br><br> dunīāṃ vicc ruzgār nāl sabndhit sabh tōṃ vaḍḍī sakīm mahātmā gāndhī rāshṭarī pēṇḍū ruzgār sakīm ("manrēgā") pañjāb vicc siāsī atē prashāsanik bēruk̲h̲ī dā shikār dikhāī dindī hai. manrēgā kānūnn nāl ġarīb parivārāṃ nūṃ sau din dā ruzgār prāpat karan dā saṃvidhānak hakk mil giā sī. ih pahilī sakīm hai jō maṅg uttē ādhārit hai, kisē sarkār jāṃ adhikārīāṃ dī garāṇṭ uttē nahīṃ. |
| Sample Output | ਦੁਨੀਆਂ,ਰੁਜ਼ਗਾਰ,ਸਕੀਮ,ਮਹਾਤਮਾ,ਗਾਂਧੀ,ਰਾਸ਼ਟਰੀ,ਪੇਂਡੂ,ਪੰਜਾਬ,ਸਿਆਸੀ,ਕਾਨੂੰਨ,ਦਿਨ,ਪ੍ਰਾਪਤ,ਸੰਵਿਧਾਨਕ,ਹੱਕ,ਮੰਗ,ਸਰਕਾਰ,ਮਨਰੇਗਾ <br><br> dunīāṃ,ruzgār,sakīm,mahātmā,gāndhī,rāshṭarī,pēṇḍū,pañjāb, siāsī,kānūnn,din,prāpat,saṃvidhānak, hakk,maṅg,sarkār, manrēgā, |

## IV. RESULTS AND DISCUSSIONS

The Punjabi text document corpus contains 1000 documents with an average of 700 Punjabi words in each document. The text documents belong to five different domains, which are Agriculture, Entertainment, Health, Politics, and Sports. Each domain contains 200 Punjabi documents. These five domains are labeled as classes for the domain ontology learning process. The documents of the corpus in every class are further divided to permit 70% of them for the learning set and 30% for the testing set. These documents are stored in text files.

Files and arrays are used as data structure for preprocessing of Punjabi text documents. Dictionary, Stop word list, and gazetteer list is kept in text files. Throughout the implementation, these lists are kept in arrays to retrieve the data quickly; otherwise, accessing contents directly from the files increase processing time.

### A. Result of Removal of duplicate terms

The analysis has been done on 30% Punjabi documents of data set for finding the frequency of duplicate terms. It is found that 56.1% duplicate terms are removed from agriculture domain documents, 52% from Entertainment domain, 57.4% from Health domain, 54% from Politics domain and 53.5% from Sports domain. The average frequency of percentage of duplicate terms in Punjabi documents is 54.6%.

Following figures show the result of removal of duplicate terms from all domains.

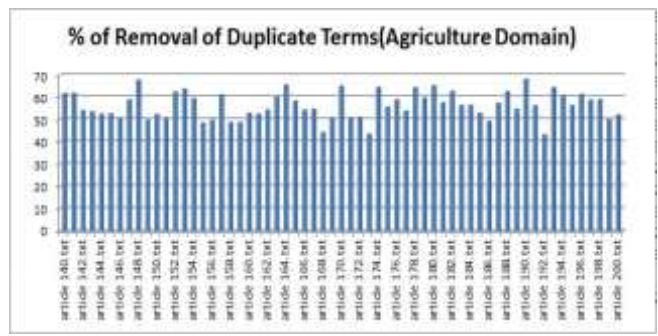

Fig. 2  Percentage of duplicate terms removed from Agriculture domain

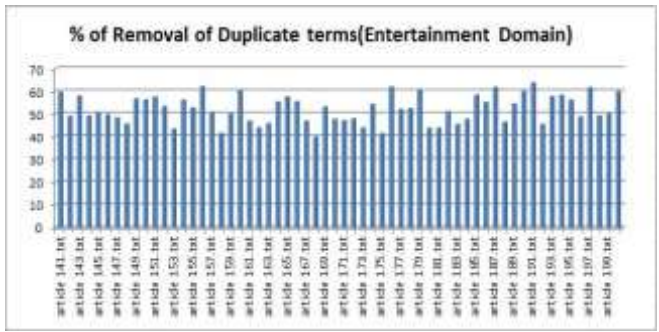

Fig. 3 Percentage of duplicate terms removed from Entertainment domain

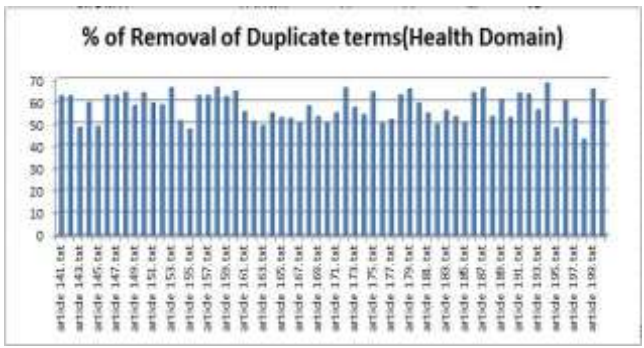

Fig. 4 Percentage of duplicate terms removed from Health domain

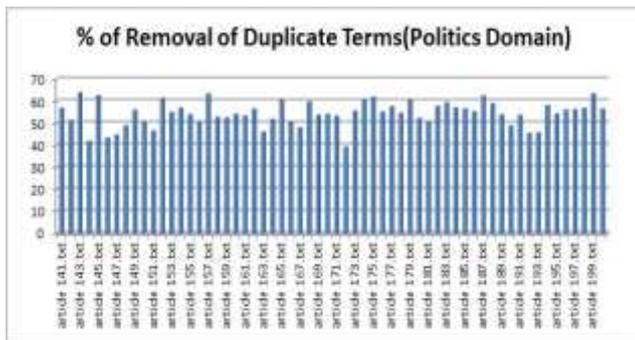

Fig. 5 Percentage of duplicate terms removed from Politics domain

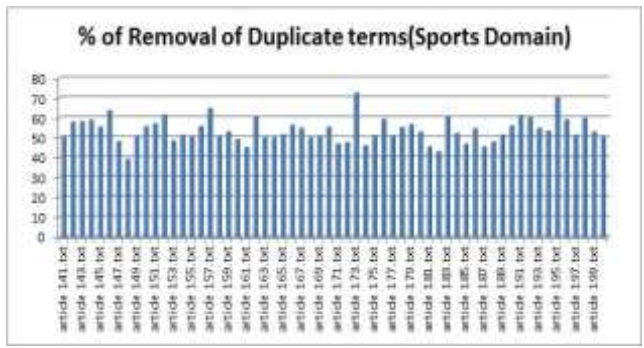

Fig. 6 Percentage of duplicate terms removed from Sports domain

### B. Result of Removal of Stop Words

The average frequency of stop word removed from each document of five domains is 98. The percentage of stop words removed from five different domains is as follows:

The documents of Agriculture Domain contains 34.44% stop words, Entertainment domain documents contains 32.7% stop words, Health domain documents contains 27% stop words, Politics domain contains 30.45% stop words and Sports domain contains 33.62% stop words.

The average of percentage of stop word removed from all domains is 34% that means 34% of the text documents are eliminated as stop words.

The following figures show the percentage of stop word removed from documents of five domains.

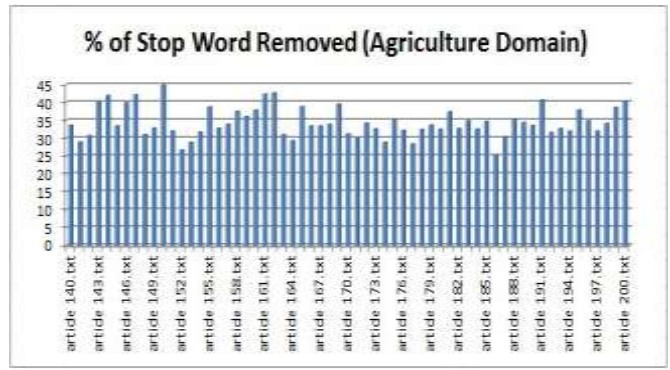

Fig. 7 Percentage of stop word removed from Agriculture domain documents.

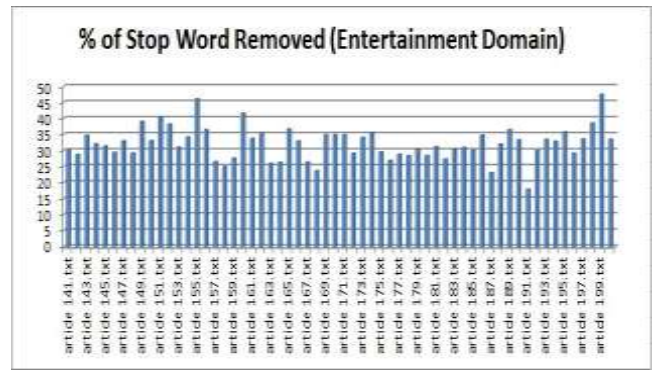

Fig. 8 Percentage of stop word removed from Entertainment domain documents.

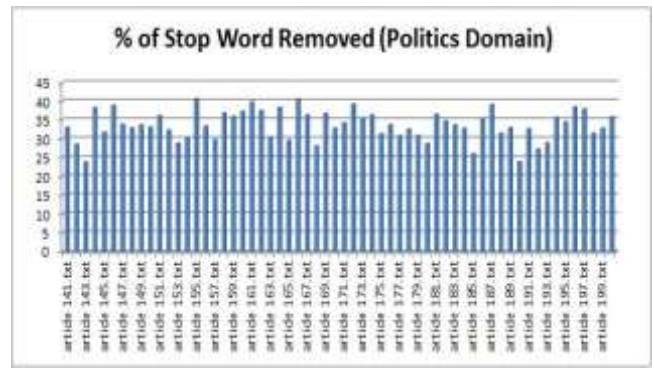

Fig. 9 Percentage of stop word removed from Politics domain documents.

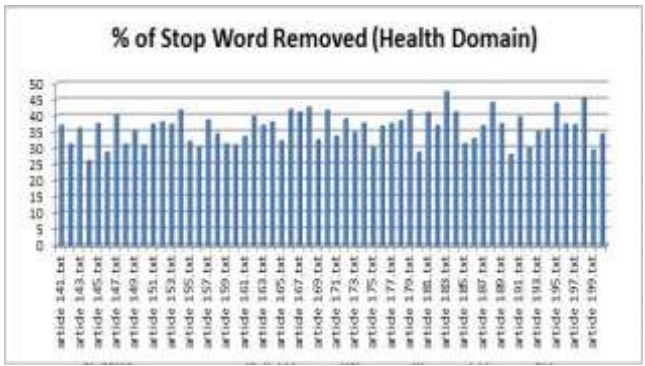

Fig. 10 Percentage of stop word removed from Health domain documents.

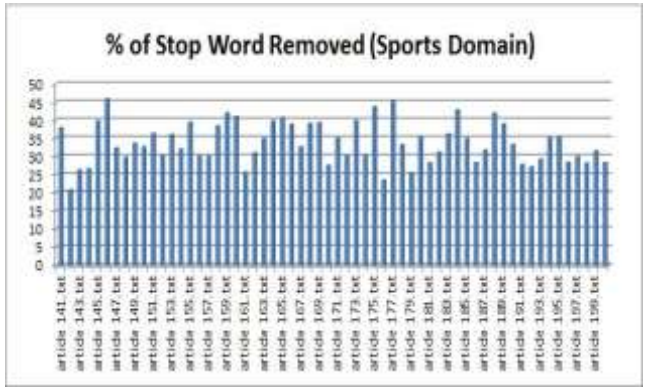

Fig. 11 Percentage of stop word removed from Sports domain documents.

*C. Result of Dictionary and Gazetteer list matched Terms*

The percentage of total matched terms (i.e. dictionary terms and gazetteer list terms) of five domains is as follows:

The percentage of total matched terms of agriculture domain is 67.3%, Entertainment domain is 66%, Health domain is 63.4%, Politics domain is 68% and Sports domain is 72%. The average percentage of total matched terms of all domains articles is 67.34%. Following figures shows the percentage of total matched terms of five different domains articles.

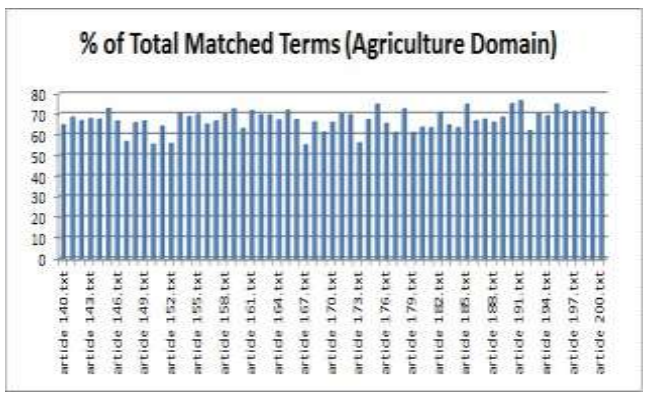

Fig. 12 Percentage of total matched terms of Agriculture domain articles.

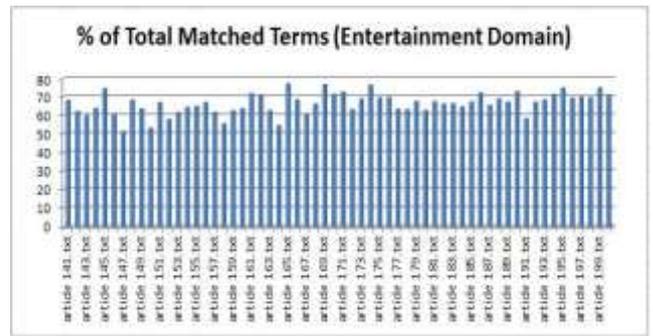

Fig. 13 Percentage of total matched terms of Entertainment domain articles.

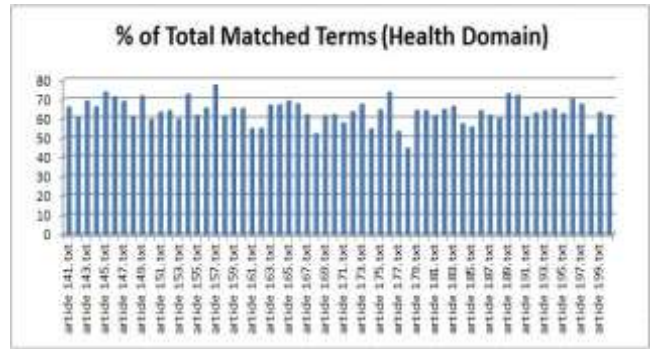

Fig. 14 Percentage of total matched terms of Health domain articles.

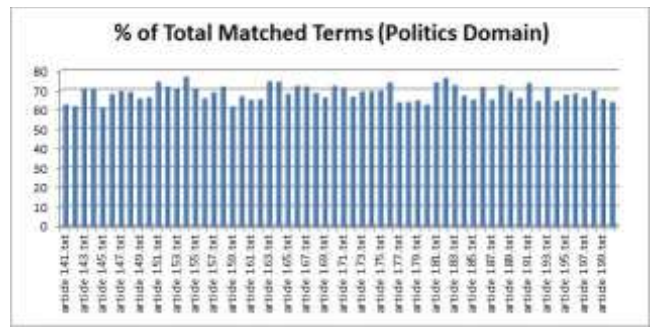

Fig. 15 Percentage of total matched terms of Politics domain articles.

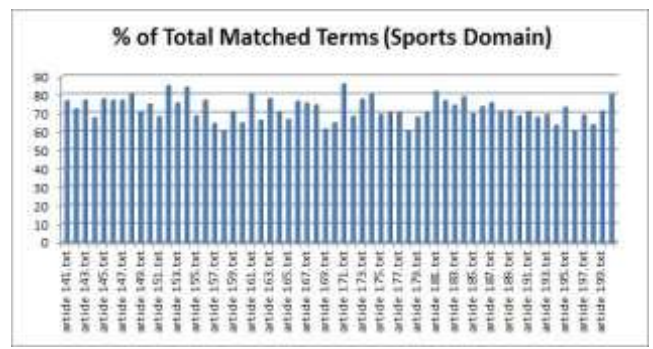

Fig. 16 Percentage of total matched terms of Sports domain articles.

*D. Final result of pre-processing*

The pre-processing has been done on 30% Punjabi documents of the data set. As shown in figure, it is discovered that 57% of total input terms are removed as duplicate terms, 14% of total input terms are stop words, 17% terms are matched with dictionary terms and 3% terms are matched with gazetteer list terms. It means 20% terms are matched with dictionary and Gazetteer list terms that are used for further processing (i.e. for processing phase).

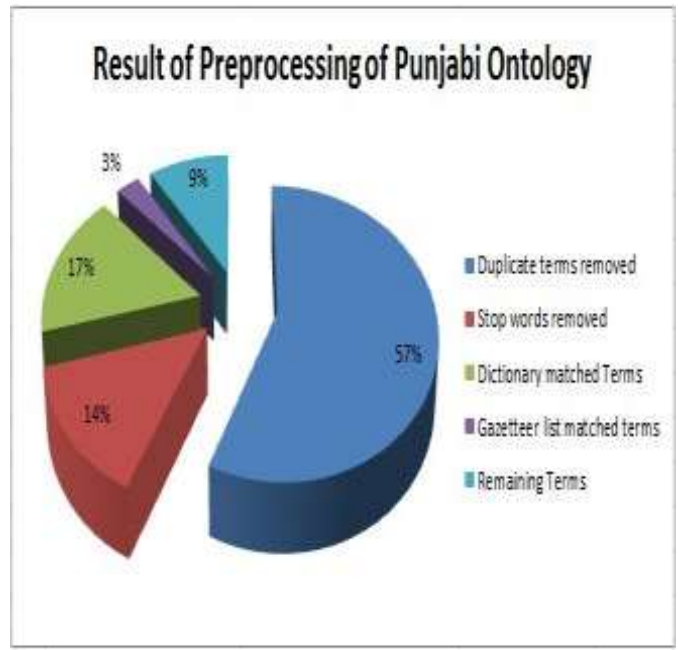

Fig. 17 Result of Pre-processing of Punjabi Ontology

The result shows that only 9% terms of total input terms are remaining terms that are meaningless and does not use for further processing, as shown in Fig. 17.

## V. CONCLUSION

In this paper, we have discussed the Pre-processing phase of Ontology graph generation system in Punjabi. The Pre-processing includes use of lexical resources such as Punjabi stop word list and gazetteer list. These lexical resources are developed manually by analyzing the Punjabi corpus as there are no Punjabi resources available on the web. These lexical resources can be helpful for developing new NLP systems in Punjabi language.

### ACKNOWLEDGMENT

I am grateful to Er. Saurabh Sharma, Assistant Professor, Baddi University of Emerging Science and Technology, Baddi (H.P.) for being my mentor and providing guidance and encouragement for this research work. His inputs made this herculean task easy for me to perform.